# Convolutional versus Self-Organized Operational Neural Networks for Real-World Blind Image Denoising

Junaid Malik[a,*], Serkan Kiranyaz[b], Mehmet Yamac[c], Esin Guldogan[c], Moncef Gabbouj[a]

[a]Department of Computing Sciences, Tampere University, Tampere, Finland.
[b]Department of Electrical Engineering, Qatar University, Doha, Qatar.
[c]HUAWEI Technologies, Tampere, Finland.

*Abstract*— **Real-world blind denoising poses a unique image restoration challenge due to the non-deterministic nature of the underlying noise distribution. Prevalent discriminative networks trained on synthetic noise models have been shown to generalize poorly to real-world noisy images. While curating real-world noisy images and improving ground truth estimation procedures remain key points of interest, a potential research direction is to explore extensions to the widely used convolutional neuron model to enable better generalization with fewer data and lower network complexity, as opposed to simply using deeper Convolutional Neural Networks (CNNs). Operational Neural Networks (ONNs) and their recent variant, Self-organized ONNs (Self-ONNs), propose to embed enhanced non-linearity into the neuron model and have been shown to outperform CNNs across a variety of regression tasks. However, all such comparisons have been made for compact networks and the efficacy of deploying operational layers as a drop-in replacement for convolutional layers in contemporary deep architectures remains to be seen. In this work, we tackle the real-world blind image denoising problem by employing, for the first time, a deep Self-ONN. Extensive quantitative and qualitative evaluations with several evaluation metrics over the four high-resolution, real-world noisy image datasets against a deep CNN network, DnCNN, reveal that deep Self-ONNs consistently achieve superior results with performance gains of up to 1.76dB in PSNR. Furthermore, Self-ONNs with half and even quarter the number of layers and neurons with significantly less computational complexity can still achieve similar or usually better results compared to DnCNN.**

## I. INTRODUCTION

Image denoising is one of the key low-level image restoration tasks because noise is inherent to every image acquisition system. A foremost challenge in designing denoising methods is the fact that obtaining a true noise-free image is an ill-posed problem because the noise in real-world images is non-deterministic. One of the most widely accepted ways to deal with this is through the use of synthetic noise models, where noise belonging to a known distribution (generally white Gaussian) is added to the original image to produce its noisy counterpart [1–7]. While this approach has been a standard for decades, it makes strong assumptions that are inconsistent with empirical observations about real-world noisy images [8–10]. Most importantly, noise in real-world images does not necessarily follow the normal distribution, is data dependent and not stationary. Moreover, the additive approach for synthesizing noisy images assumes the original image to be free of any kind of noise, which is rarely the case. Because of this, discriminative learning-based denoising methods that are trained on synthetic noisy images fail to maintain their performance levels when tested on real-world images [8,11]. To remedy this, several image datasets have been proposed in recent years, which employ different ground truth estimation procedures to produce clean images from real-world noisy images [8,12,13]. Discriminative methods trained on these new datasets have exhibited a better generalization performance than those trained using synthetic additive noise data [8]. While the efforts for producing better estimations for ground-truth images continue, it is worth noting that the key ingredients behind the denoising methods have not yet progressed.

Almost all discriminative learning-based methods utilize Convolutional Neural Networks (CNNs) [4,7,14], which have become a *de-facto* standard for many computer vision tasks. However, recent studies have identified drawbacks in the convolutional neuron model [15], especially for denoising [16]. Being a variant of the multi-layer perceptron (MLP) with two known restrictions (i.e., limited connection and weight sharing), convolutional neurons essentially apply a linear transformation to the input (i.e., the output map of the previous layer neuron), whereas the point-wise activation function or non-linear pooling operations following the convolution are the unique sources of non-linearity in the network. This is often cited as one of the major reasons why *state-of-the-art* CNNs need to be deep, over-parameterized, and hence require a large amount of training data to achieve a decent generalization. Many possible solutions have been explored recently [17,18] to increase non-linearity in CNNs without adding more layers. One of the key developments in this regard is the conception of so-called Operational Neural Networks (ONNs) [15]. ONNs are composed of generalized (operational) neurons that allow the flexibility to incorporate any non-linear patch-wise transformation. When properly configured with optimal non-linear functions, ONNs were shown to outperform equivalent CNNs across a variety of challenging tasks and under severe training constraints. More importantly, unlike [17,18], ONNs allow a heterogeneous network model where each neuron applies a unique transformation to the input, resulting in enhanced diversity of intra-layer activations. An important factor that determines the performance of ONNs is the choice of non-linear operators. There have been numerous search methodologies explored, including GIS [15] and synaptic plasticity monitoring (SPM) [19] to identify the best operators for a given learning problem. However, they require additional training runs to ensure convergence towards suitable operators. To address this drawback, a self-organized variant of ONNs, the Self-ONN, has been proposed recently [16,20]. Self-ONNs employ the Taylor series function approximation to generate non-linear transformations during the training. Self-ONNs were successfully applied for a variety of severe image restoration problems in [16,21]. Compact (two-layer) Self-ONNs outperformed CNNs across diverse noise models with a significant gap. The work established, based on various ablation tests, that with limited training data and compact networks, Self-ONNs can learn and generalize better than CNNs. However, the value of the proposed Self-ONNs for deeper network architectures and more realistic noise characteristics remains to be seen.

* Corresponding Author
 Email: junaid.malik@tuni.fi

In this study, we investigate the efficacy of deep Self-ONNs for denoising real-world noisy images, as compared to CNNs. To this end, we consider a 17-layer deep denoising architecture of DnCNN [4] and convert it to a Self-ONN by replacing convolutional layers with the (self-organized) operational layers. All networks are subsequently trained with the same training data and equivalent training conditions to highlight the true impact of introducing the operational layers. The primary contributions of this work are enumerated as follows:

- We configure and train a novel deep Self-ONN for real-world image denoising. This is the first study that proposes a deep ONN for a regression task. This is also the first study that proposes an ONN for real-world denoising problems.
- We evaluate and compare Self-ONN with DnCNN across 4 high-resolution real-world denoising benchmark data and demonstrate that a Self-ONN, even with a fewer number of layers, surpasses DnCNN, especially in terms of perceptual quality metrics.
- We analyze and discuss the effect of enhanced nonlinearity and heterogeneity in deep Self-ONNs quantitatively by calculating the strengths of the synaptic connections.
- We provide an alternative vectorized formulation of Self-ONNs which lends itself to efficient parallelization on dedicated hardware resources such as GPUs.

The remaining of the paper is organized as follows: Section II briefly presents the related work including the benchmark datasets in real-world image denoising. Section III briefly reviews Self-ONNs and generative neurons and formulates the novel vectorized notation. Section IV introduces the experimental setup and presents detailed comparative evaluations of Self-ONNs and DnCNN over four benchmark datasets and discusses the results. Moreover, the computational complexity analysis of the proposed networks is also presented in this section. Finally, Section V concludes the paper and suggests topics for future research.

## II. RELATED WORK

### A. Datasets

The RENOIR dataset [13] consists of images from two digital cameras and one smartphone. It contains 120 images corresponding to as many scenes. Noisy images were acquired by using a short-time exposure camera setting to capture low-light scenes. The corresponding reference images are acquired by long-time exposure to the same scene. Intensity alignment is applied as a post-processing step to produce "ground-truth" images. The CCNoise dataset [22] contains images from 11 natural scenes captured through 3 cameras. It consists of 17 images acquired by taking 500 consecutive images taken from the same scene. Assuming that the noise is zero-mean, ground truth estimation is done by taking the temporal mean of these 500 images. The DND dataset [8] consists of images from 50 unique scenes. Reference/clean image is acquired by capturing the scene using a base ISO level. The corresponding noisy image is taken using a higher ISO/lower exposure time. For ground truth estimation, a thorough post-processing procedure consisting of removal of low-frequency residual noise and intensity alignment is used. For benchmarking, 512x512 test crops are provided whereas the ground truth images are not made public and are held out for benchmarking through online submission. SIDD dataset [12] was curated exclusively for smartphone images by capturing 10 different scenes with 15 different ISO values (corresponding to different levels of noise). Images were acquired using five different smartphones with different illumination temperatures and brightness levels, leading to a vast variety of noisy images. In addition, an extensive ground-truth estimation procedure was adopted to obtain clean images. The "MEDIUM" variant of the dataset consists of 320 high-resolution noisy-clean image pairs.

### B. Real-World Denoising Methods

Numerous denoising techniques were benchmarked on the RENOIR dataset in [13]. The pioneering non-local technique, BM3D [23], achieved considerably superior results compared to all other techniques, providing an appropriate estimate of the noise level. In particular, the discriminative method of MLP [3], which was proven to be competitive with BM3D on synthetic noise, could not retain the performance on real-world noisy images. Its performance fell on average by 2.06dB in terms of PSNR and around 5% in terms of the perceptual quality index (SSIM). Benchmarking on later datasets [8,12] confirmed these findings, despite employing vastly different ground-truth estimation methods. Even sophisticated deep CNN-based methods such as DnCNN [4] trained on synthetic AWGN noise models did not provide adequate denoising performance. Nevertheless, the same architecture performed considerably better when trained on real-world instead of synthetic noisy-clean image pairs. Considering these findings, contemporary methods generally aim at utilizing available real-world noisy images [24–28], devising realistic noise models to synthesize artificial noisy images [11,26,29,30], and designing appropriate architectural changes [11,24–28,30–33]. It is worth noting that the common denominator among all these networks is the sole usage of the convolutional layer as the core transformation. Their novelty is limited to architectural upgrades such as residual connections [28], feature attention [25], multi-scale approach [30], or some combination of these techniques. In contrast, in this study, while keeping the network architecture of DnCNN intact, we propose a fundamental paradigm shift by proposing the operational layers of Self-ONNs instead of the convolutional layers to enhance the expressivity of the denoising models. To validate our claim, we employ the deep CNN architecture of DnCNN [4] and convert it to a Self-ONN with the drop-in replacement of all convolutional layers. To expel the subjectivity arising from the choice of training data and hyperparameters, all the networks in this study are identically trained from scratch using the same training resources. We show that Self-ONNs can significantly surpass DnCNN and remain competitive even when trimmed down to four-times fewer layers, especially in terms of perceptual quality of the denoised images.

## III. SELF-ORGANIZED OPERATIONAL NEURAL NETWORKS

In this section, we first remind the reader how ONNs [15] replaces the convolution operation with any nonlinear nodal operation, then briefly review the generative neuron-based Self-ONN [16,20] which is the basis of the proposed network. Finally, a novel vectorized implementation of Self-ONNs will be introduced.

We consider the case of the $k^{th}$ neuron in the $l^{th}$ layer of a 2D convolutional neural network. The output of this neuron is formalized as follows:

$$x_k^l = b_k^l + \sum_{i=0}^{N_{l-1}} x_{ik}^l \quad (1)$$

where $b_k^l$ is the bias associated with this neuron and $x_{ik}^l$ is defined as in (2).

$$x_{ik}^l = Conv2D(w_{ik}, y_i^{l-1}) \tag{2}$$

where $Conv2D$ refers to the 2D convolution operation, $w_{ik} \in \mathbb{R}^{K \times K}$ is the kernel connecting the $i^{th}$ neuron of $(l-1)^{th}$ layer to $k^{th}$ neuron of $l^{th}$ layer, while the input $x_{ik}^l \in \mathbb{R}^{M \times N}$ and the output $y_i^{l-1} \in \mathbb{R}^{M \times N}$ correspond to the $k^{th}$ and $i^{th}$ neurons' outputs of the $l^{th}$ and $(l-1)^{th}$ layers, respectively. By definition, the convolution operation of (2) can be expressed as follows:

$$x_{ik}^l(m,n) = \sum_{r=0}^{K-1} \sum_{t=0}^{K-1} w_{ik}^l(r,t) y_i^{l-1}(m+r, n+t) \tag{3}$$

The operational neuron [15] generalizes the convolution operation in CNN as follows:

$$\overline{x_{ik}^l}(m,n) = P_k^l \left( \psi_k^l \left( w_{ik}^l(r,t), y_i^{l-1}(m+r, n+t) \right) \right)_{(r,t)=(0,0)}^{(K-1,K-1)} \tag{4}$$

where $\psi_l^k(\cdot): \mathbb{R}^{MN \times K^2} \to \mathbb{R}^{MN \times K^2}$ and $P_k^l(\cdot): \mathbb{R}^{MN \times K^2} \to \mathbb{R}^{MN}$ are called *nodal* and *pool* operators, respectively, and assigned to the $k^{th}$ neuron of $l^{th}$ layer. In a heterogenous ONN configuration, every neuron has uniquely assigned $\psi$ and $P$ operators. To resolve ONN challenges with respect to the selection of suitable operators (see [15] for more details) Self-organized Neural Networks (Self-ONNs) [16,20] introduced a composite nodal function that is iteratively created and tuned during back-propagation using the Taylor-series-based function approximation. The Taylor series expansion of an infinitely differentiable function $f(x)$ about a point $a$ is given as:

$$f(x) = \sum_{n=0}^{\infty} \frac{f^{(n)}(a)}{n!} (x-a)^n \tag{5}$$

The $Q^{th}$ order truncated approximation of (5), formally known as the Taylor polynomial, takes the following form:

$$f(x)^{(Q,a)} = \sum_{n=0}^{Q} \frac{f^{(n)}(a)}{n!} x^n \tag{6}$$

The above formulation enables the approximation of any function $f(x)$ sufficiently well in the close vicinity of $a$. If the coefficients $\frac{f^{(n)}}{n!}$ are tuned and the inputs are bounded, the formulation of (6) can be used to *generate* any *nodal* operator. This is the key idea behind the *generative* neurons of a Self-ONN [20]. Following the notation used in (4), the nodal operator of a generative neuron takes the general form of (7).

$$\widetilde{\psi_k^l}\left(w_{ik}^{l(Q)}(r,t), y_i^{l-1}(m+r, n+t)\right) = \sum_{q=1}^{Q} w_{ik}^{l(Q)}(r,t,q) \left(y_i^{l-1}(m+r, n+t)\right)^q \quad (7)$$

where $Q$ is a hyperparameter that controls the degree of the Taylor approximation, and $w_{ik}^{l(Q)}$ is a learnable kernel of the network. A key difference in (7) as compared to the convolution operation in (3) and the operational model in (4) is that $\widetilde{\psi_k^l}$ is not fixed a priori; instead, it is a distinct operator over each individual output, $y_i^{l-1}$, and thus requires $Q$ times more parameters. Therefore, the $K \times 1$ kernel vector $w_{ik}^l$ is replaced by a $K \times Q$ matrix $w_{ik}^{l(Q)} \in \mathbb{R}^{K \times Q}$ which is formed by replacing each element $w_{ik}^l(r)$ with a $Q$-dimensional vector $w_{ik}^{l(Q)}(r) = [w_{ik}^{l(Q)}(r,0), w_{ik}^{l(Q)}(r,1), \dots, w_{ik}^{l(Q)}(Q-1)]$. The input map of the generative neuron, $\tilde{x}_{ik}^l$ can now be expressed as,

$$\widetilde{x_{ik}^l}(m,n) = P_k^l \left( \sum_{q=1}^{Q} w_{ik}^{l(Q)}(r,t,q) \left(y_i^{l-1}(m+r, n+t)\right)^q \right)_{(r,t)=(0,0)}^{(K-1, K-1)} \quad (8)$$

Recall that during training, as $w_{ik}^{l(Q)}$ is iteratively tuned by back-propagation (BP), *customized* nodal transformation functions are generated as a result of (8), which would be tailored for $i-k^{th}$ connection. Moreover, unlike ONNs, the generative neurons of a Self-ONN layer can be parallelized more efficiently, leading to a considerable reduction in computational complexity and time as will be shown later. Finally, a specific case of formulation (8) can be expressed in terms of the widely used convolutional model, as will be shown next.

## A. Representation of Self-ONN Using Convolutions

Consider the pooling operator $P_k^l$ to be the summation operator. $\tilde{x}_{ik}^l$ is then defined as:

$$\widetilde{x_{ik}^l}(m) = \sum_{r=0}^{K-1} \sum_{t=0}^{K-1} \sum_{q=1}^{Q} w_{ik}^{l(Q)}(r,t,q) \left(y_i^{l-1}(m+r, n+t)\right)^q \quad (9)$$

Exploiting the commutativity of the summation operations in (9), we can alternatively write:

$$\widetilde{x_{ik}^l}(m) = \sum_{q=1}^{Q} \sum_{r=0}^{K-1} \sum_{t=0}^{K-1} w_{ik}^{l(Q)}(r,t,q \\ -1) y_i^{l-1}(m+r, n+t)^q \quad (10)$$

Using (1) and (2), the formula in (10) can be further simplified as in (11).

$$\widetilde{x_{ik}^l} = \sum_{q=1}^{Q} Conv2D\left(w_{ik}^{l(Q)}, \left(y_i^{l-1}\right)^q\right) \quad (11)$$

Hence, the formulation can be accomplished by applying $Q$ 2D convolution operations. If $Q$ is set to 1, (11) entails the convolutional formulation of (3). Therefore, as CNN is a subset of ONN corresponding to a specific operator set, it is also a special case of a Self-ONN in which $Q = 1$ set for all neurons.

B. *Vectorized Notation*

Expressing explicit loops in terms of matrix and vector manipulations is the key idea behind vectorization, which enabled fast implementations of modern-day neural network implementations. As noted in [16], the $Q$ terms in the summation of (11) can be calculated independently, and thus lend themselves to parallelization. In this section, we introduce a novel formulation that exploits this property and enables the formulation of (11) to be achieved using a single matrix-vector product. We first introduce how a vectorized notation can be used to express the 2D convolution operation inside a neuron. Afterward, the same key principles will be exploited to express the generative neuron formulation of (9) as a single matrix-vector product.

An alternate formulation of the operation of (3) is now presented. We introduce a transformation $\delta(\cdot, K^2)$ using which $y_i^{l-1}$ is reshuffled such that values inside each $K \times K$- kernel across $y_i^{l-1}$ are vectorized and concatenated as rows to form a matrix $Y_i^{l-1} \in \mathbb{R}^{M \times N \times K}$. The process is visually depicted in Fig. 7 for $K = 2$, and mathematically expressed in (12).

$$Y_i^{l-1} = \delta\left(y_i^{l-1}, K^2\right) = \begin{bmatrix} \cdots & y_i^{l-1}(m,n) & \cdots \\ \vdots & \vdots & \vdots \\ \cdots & y_i^{l-1}(m, n+K-1) & \cdots \\ \cdots & y_i^{l-1}(m+1, n) & \cdots \\ \vdots & \vdots & \vdots \\ \cdots & y_i^{l-1}(m+1, n+K-1) & \cdots \\ \vdots & \vdots & \vdots \\ \cdots & y_i^{l-1}(m+K-1, n) & \cdots \\ \vdots & \vdots & \vdots \\ \cdots & y_i^{l-1}(m+K-1, n+K-1) & \cdots \end{bmatrix}^T \quad (12)$$

Secondly, we construct a matrix $W_{ik}^l \in \mathbb{R}^{M \times K}$ whose rows are repeated copies of the vectorized version of $w_{ik} \in \mathbb{R}^{K \times K}$.

$$W_{ik}^l = \begin{bmatrix} w_{ik}^l(0,0) & w_{ik}^l(0, K-1) & \cdots & w_{ik}^l(K-1, K-1) \\ \vdots & \vdots & \cdots & \vdots \\ w_{ik}^l(0,0) & w_{ik}^l(1, K-1) & \cdots & w_{ik}^l(K-1, K-1) \\ \vdots & \vdots & \cdots & \vdots \\ w_{ik}^l(0,0) & w_{ik}^l(1, K-1) & \cdots & w_{ik}^l(K-1, K-1) \end{bmatrix} \quad (13)$$

We now consider the Hadamard product of these two matrices:

$$Y_i^{l-1} \otimes W_{ik}^l = \begin{bmatrix} y_i^{l-1}(0,0)w_{ik}^l(0,0) & \cdots & y_i^{l-1}(K-1,K-1)w_{ik}^l(K-1,K-1) \\ \vdots & \cdots & \vdots \\ y_i^{l-1}(m,n)w_{ik}^l(0,0) & \cdots & y_i^{l-1}(m+K-1,n+K-1)w_{ik}^l(K-1,K-1) \\ \vdots & \cdots & \vdots \\ y_i^{l-1}(M-1,N-1)w_{ik}^l(0,0) & \cdots & y_i^{l-1}(M+K-1,N+K-1)w_{ik}^l(K-1,K-1) \end{bmatrix} \quad (14)$$

Applying summation operation across rows, we get:

$$\sum (Y_i^{l-1} \otimes W_{ik}^l) = \begin{bmatrix} \sum_{r=0}^{K-1}\sum_{t=0}^{K-1} y_i^{l-1}(r,t)w_{ik}^l(r,t) \\ \vdots \\ \sum_{r=0}^{K-1}\sum_{t=0}^{K-1} y_i^{l-1}(m+r,n+t)w_{ik}^l(r,t) \\ \vdots \\ \sum_{r=0}^{K-1}\sum_{t=0}^{K-1} y_i^{l-1}(M+r,N+t)w_{ik}^l(r,t) \end{bmatrix} = \sum_{r=0}^{K-1} w_{ik}^l(r)y_i^{l-1}(m+r) \quad (15)$$

Reshaping this back to $\mathbb{R}^{M \times N}$, we get

$$vec_{M \times N}^{-1}\left(\sum(Y_i^{l-1} \otimes W_{ik}^l)\right)(m,n) = \sum_{r=0}^{K-1}\sum_{t=0}^{K-1} y_i^{l-1}(m+r,n+t)w_{ik}^l(r,t) \quad (16)$$

which is equivalent to (3). We also note that,

$$\sum Y_i^{l-1} \otimes W_{ik}^l = Y_i^{l-1}\overrightarrow{w_{ik}^l} \quad (17)$$

Therefore,

$$x_{ik}^l(m,n) = vec_{M \times N}^{-1}\left(Y_i^{l-1}\overrightarrow{w_{ik}^l}\right)(m,n) \quad (18)$$

$$x_{ik}^l = vec_{M \times N}^{-1}\left(Y_i^{l-1}\overrightarrow{w_{ik}^l}\right) \quad (19)$$

We show here a representation of 2D convolution operation to compute the output of a neuron by a single matrix-vector product. This operation lies at the heart of conventional generalized matrix multiplication (GEMM)-based convolution implementations and enables efficient usage of parallel computational resources such as GPU cores.

*C. Forward Propagation through a 2D Self-ONN neuron*

We showed in (11) how the Self-ONN formulation of (10) can be represented as a summation of $Q$ individual convolutional operations. Moreover, from (12), a convolutional operation can be represented as a matrix-vector product. We now use these two formulations to represent the transformation of (11) as a single convolution operation, and consequently a single matrix-vector product, instead of Q-separate ones. We start by introducing $Y_i^{l-1^{(Q)}} \in \mathbb{R}^{MN \times K^2 Q}$ such that

$$Y_i^{l-1^{(Q)}} = \begin{bmatrix} Y_i^{l-1} & \left(Y_i^{l-1}\right)^{\circ 2} & \cdots & \left(Y_i^{l-1}\right)^{\circ Q} \end{bmatrix} \quad (20)$$

where $\circ n$ is the Hadamard exponentiation operator. The $m^{th}$ row of $Y_i^{l-1^{(Q)}}$ can be expressed as

$$Y_i^{l-1^{(Q)}} = \begin{bmatrix} \cdots & y_i^{l-1}(m,n) & \cdots \\ \vdots & \vdots & \vdots \\ \cdots & y_i^{l-1}(m,n+K-1) & \cdots \\ \vdots & \vdots & \vdots \\ \cdots & y_i^{l-1}(m+K-1,n+K-1) & \cdots \\ \vdots & \vdots & \vdots \\ \cdots & y_i^{l-1}(m,n)^2 & \cdots \\ \vdots & \vdots & \vdots \\ \cdots & y_i^{l-1}(m,n+K-1)^2 & \cdots \\ \vdots & \vdots & \vdots \\ \cdots & y_i^{l-1}(m+K-1,n+K-1)^2 & \cdots \\ \vdots & \vdots & \vdots \\ \cdots & y_i^{l-1}(m,n)^Q & \cdots \\ \vdots & \vdots & \vdots \\ \cdots & y_i^{l-1}(m,n+K-1)^Q & \cdots \\ \vdots & \vdots & \vdots \\ \cdots & y_i^{l-1}(m+K-1,n+K-1)^Q & \cdots \end{bmatrix}^T \quad (21)$$

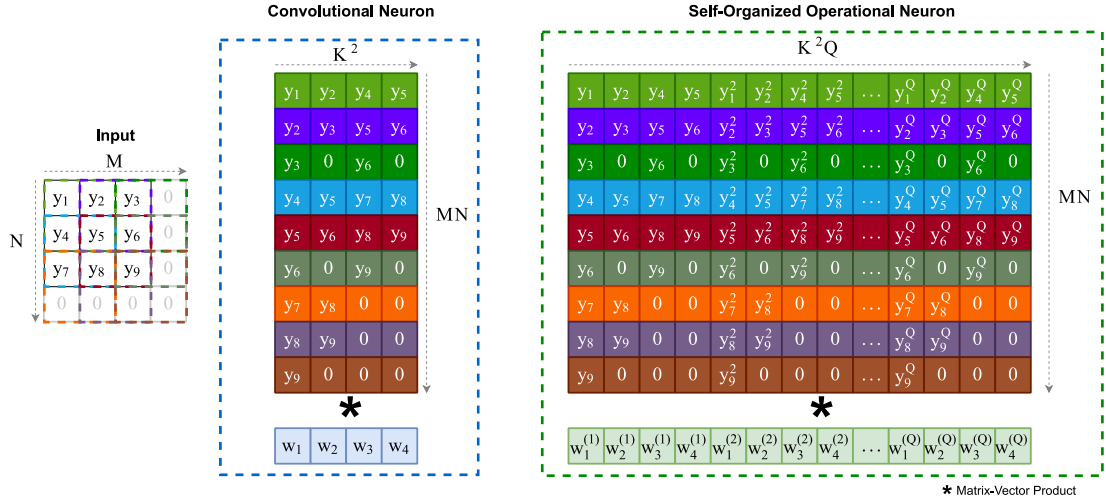

*Figure 1. A visual illustration of the transformations required for convolutional and self-organized operational neurons. Both neuron models require a single matrix-vector product.*

This is also illustrated in Figure 1. Moreover, we construct $W_{ik}^{l(Q)} \in \mathbb{R}^{MN \times KQ}$ by first vectorizing $w_{ik}^{l(Q)} \in \mathbb{R}^{K \times K \times Q}$ to $\overrightarrow{w_{ik}^{l(Q)}} \in \mathbb{R}^{K^2 Q}$ and then concatenating $m$ copies of $\overrightarrow{w_{ik}^{l(Q)}}$ along the row dimension, as expressed in (22) and (23).

$$\overrightarrow{w_{ik}^{l(Q)}} = \begin{bmatrix} w_{ik}^{l(Q)}(0,0,0) \\ \vdots \\ w_{ik}^{l(Q)}(K-1,K-1,0) \\ w_{ik}^{l(Q)}(0,0,1) \\ \vdots \\ w_{ik}^{l(Q)}(K-1,K-1,1) \\ \vdots \\ w_{ik}^{l(Q)}(0,0,Q-1) \\ \vdots \\ w_{ik}^{l(Q)}(K-1,K-1,Q-1) \end{bmatrix}^T \quad (22)$$

$$W_{ik}^{l(Q)}(m) = \overrightarrow{w_{ik}^{l(Q)}} \quad (23)$$

The matrix-vector product of $Y_i^{l-1(Q)}$ and $\overrightarrow{w_{ik}^{l(Q)}}$, is now given as:

$$Y_i^{l-1(Q)}\left(\overrightarrow{w_{ik}^{l(Q)}}\right) = \begin{bmatrix} \sum_{q=1}^{Q}\sum_{r=0}^{K-1}\sum_{t=0}^{K-1} y_i^{l-1}(r,t)w_{ik}^l(r,t,q-1) \\ \vdots \\ \sum_{q=1}^{Q}\sum_{r=0}^{K-1}\sum_{t=0}^{K-1} y_i^{l-1}(m+r,n+t)w_{ik}^l(r,t,q-1) \\ \vdots \\ \sum_{q=1}^{Q}\sum_{r=0}^{K-1}\sum_{t=0}^{K-1} y_i^{l-1}(M+r,N+t)w_{ik}^l(r,t,q-1) \end{bmatrix} \quad (24)$$

From (24), after reshaping back to $\mathbb{R}^{M \times N}$ we can note that:

$$\text{vec}_{M \times N}^{-1}\left(Y_i^{l-1(Q)}\left(\overrightarrow{w_{ik}^{l(Q)}}\right)\right)(m,n) = \sum_{q=1}^{Q}\sum_{r=0}^{K-1}\sum_{t=0}^{K-1} y_i^{l-1}(m+r,n+t)w_{ik}^l(r,t,q-1) \quad (25)$$

This is equivalent to (10). So, one can now express,

$$\widetilde{x_{ik}^l} = vec_{M \times N}^{-1}\left(Y_i^{l-1(Q)}\left(\overrightarrow{w_{ik}^{l(Q)}}\right)\right) \quad (26)$$

The formulation of (26) provides a key computational benefit, as the forward propagation through the generative neuron is accomplished using a single matrix-vector product. Hence, in theory, if the computational cost and memory requirement of constructing and storing matrices $Y_i^{l-1(Q)}$ and $W_{ik}^{l(Q)}$ are considered negligible, the complexity of a generative neuron is approximately the same as that of the convolutional neuron, as both can be accomplished by a single matrix-vector product that is fully parallelizable. Finally, to complete the forward propagation step, using (1), we can express:

$$\widetilde{x_k^l} = b_k^l + \sum_{i=0}^{N_{l-1}} \widetilde{x_{ik}^l} \qquad (27)$$

## IV. EXPERIMENTAL RESULTS

All experiments were conducted in Python using the FastONN library [34]. Access to source codes is available at request from *www.github.com/junaidmalik09/rwdenoise*.

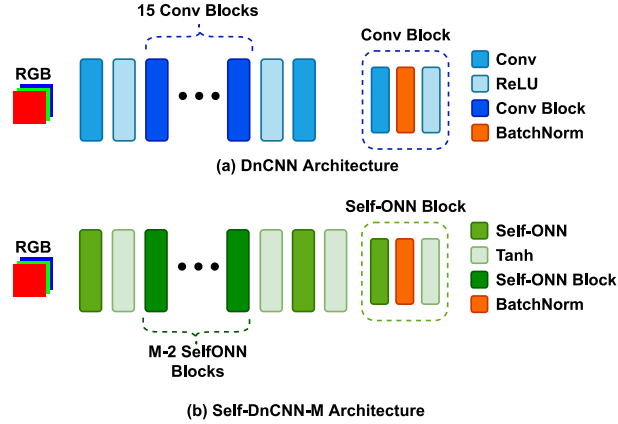

*Figure 2 Network architecture for (a) the original DnCNN* [4] *network, and (b) the Dn-SelfONN-M network.*

*A. Network Architecture*

We use DnCNN deep network [4] as the baseline CNN architecture. To keep the comparison fair between the two network models, we do not selectively convert specific convolutional layers to operational layers. Instead, we use the operational layers of Self-ONNs as a drop-in replacement for all the convolutional layers in the network. Additionally, different variants of the Self-ONN networks, Dn-SelfONN-M, are explored, each having M operational layers. For Self-ONN networks, the value of $Q$ is set to 3 by default. For Dn-SelfONN-4, a higher $Q$ value of 5 is also used. Further details are presented in Table 1.

## B. Settings

For training the network, 512 patches of size 80x80 are extracted at random from each of the 320 images of the SIDD-Medium dataset. This effectively gives a total of 160K training patches which are randomly split with a ratio of 9:1 for training and validation respectively. All the networks are trained for 500 epochs and the model achieving the best validation PSNR is chosen.

*Table 1. Details of the network architectures used in this study.*

| Network | Layers | Neurons | Q | Params (k) |
|---|---|---|---|---|
| DnCNN | 17 | 1024 | - | 558 |
| Dn-SelfONN-17 | 17 | 1024 | 3 | 1671 |
| Dn-SelfONN-8 | 8 | 512 | 3 | 675 |
| Dn-SelfONN-4_3 | 4 | 256 | 3 | 232 |
| Dn-SelfONN-4_5 | 4 | 256 | 5 | 386 |

## C. Denoising Performance

We present two sets of comparisons in this section. First, a comparison of Self-ONN networks with the baseline CNN network is presented across the datasets detailed in Section A. Afterward, we evaluate the trained models on public benchmarks provided by SIDD and DnD datasets and compare Self-ONN networks with the baseline CNN as well as contemporary works.

*Table 2. Comparison of baseline CNN with corresponding Self-ONN architectures.*

|  | RENOIR | | CC | | SIDD | |
|---|---|---|---|---|---|---|
|  | PSNR | SSIM | PSNR | SSIM | PSNR | SSIM |
| Noisy (Input) | 27.38 | 0.7459 | 34.68 | 0.9436 | 23.66 | 0.5413 |
| DnCNN (Baseline CNN) | 32.34 | 0.9034 | 36.25 | 0.9689 | 36.06 | 0.9380 |
| Dn-SelfONN-17 | **34.10**↑ | 0.9317↑ | **37.52**↑ | *0.9800*↑ | **36.95**↑ | *0.9506*↑ |
| Dn-SelfONN-8 | 33.88↑ | *0.9350*↑ | 35.20↓ | 0.9760↑ | 36.24↑ | 0.9491↑ |
| Dn-SelfONN-4_3 | 33.06↑ | 0.9227↑ | 33.83↓ | 0.9679↓ | 35.39↓ | 0.9437↑ |
| Dn-SelfONN-4_5 | 33.26↑ | 0.9261↑ | 34.85↓ | 0.9711↑ | 35.41↓ | 0.9444↑ |

*1) DnCNN vs Dn-SelfONN-17*

Experimental results for all tested models across the three datasets are shown in Table 2. One can observe that the Self-ONN model Dn-SelfONN-17 outperforms DnCNN by a substantial margin across all datasets and for both evaluation metrics. On average, improvements of 1.31 dB in terms of PSNR and 1.73% in terms of SSIM are observed across the three datasets. Hence, the deep Self-ONN network surpasses DnCNN with an equal number of layers and neurons. This establishes the superiority of the operational layers over the convolutional layers for the denoising problem.

*2) DnCNN versus Dn-SelfONN-M*

As explained in Section III, a Self-ONN layer (with Q>1) introduces extra trainable parameters as compared to a CNN. If the number of parameters is considered as the metric of equivalence, then the comparison presented in Section C might be deemed unfair. As an alternate ablation test, we reduced the number of layers and neurons in Self-ONN variants as described in Table 1 to make the comparison fair in terms of the cumulative number of trainable parameters in the network. From Table 2, we can see that the Self-ONN network with 8 layers (Dn-SelfONN-8) with half the neurons still surpasses the 17-layer CNN in all but one dataset in terms of PSNR, and all datasets in terms of SSIM metric. To take the deep CNN *versus* compact Self-ONN comparison a step further, we trimmed the Self-ONN network down to 4 layers only with a quarter of the neurons used in DnCNN. This is an unfair comparison as DnCNN is not only deeper than the Self-ONN network with four-times more learning units, but also has a significantly higher number of trainable parameters. We can see from Table 2 that both of the 4-layer Self-ONN variants (Dn-SelfONN-4_3 and Dn-SelfONN-4_5) still produce competitive results. Dn-SelfONN-4_3 surpasses the deep CNN in terms of SSIM on two datasets, and in terms of PSNR on one dataset. Specifically, improvements of 2.2% PSNR and 2.5% SSIM on the RENOIR dataset, and 0.6% SSIM on the SIDD dataset are observed, with a significant reduction of 58.5% over the trainable parameters, as compared to the baseline CNN. Increasing the value of Q to 5 improves the results further, albeit at the expense of additional trainable parameters. However, as shown in Table 1, the total number of parameters is still less than that of DnCNN so the comparison remains biased. Dn-SelfONN-4_5 outperforms the deep CNN with respect to PSNR on the RENOIR dataset and in terms of SSIM on *all* three datasets. Specifically, improvements of 2.84% PSNR and 2.5% SSIM on the RENOIR dataset, 0.22% SSIM on the CCNoise dataset, and 0.68% SSIM on the SIDD dataset are observed, with 30.8% fewer parameters than that of the baseline CNN. It is worth noting here that Self-ONNs manage to achieve a higher SSIM value than the baseline CNN, even in the cases when PSNR is lower. This can be attributed to the fact that the Self-ONN neuron model is more capable of recovering perceptual information, such as textures and local details, from the corrupted image, instead of over-smoothing which might result in better PSNR values but lower SSIM. This is further validated with the qualitative results presented in Figure 3 and Figure 4.

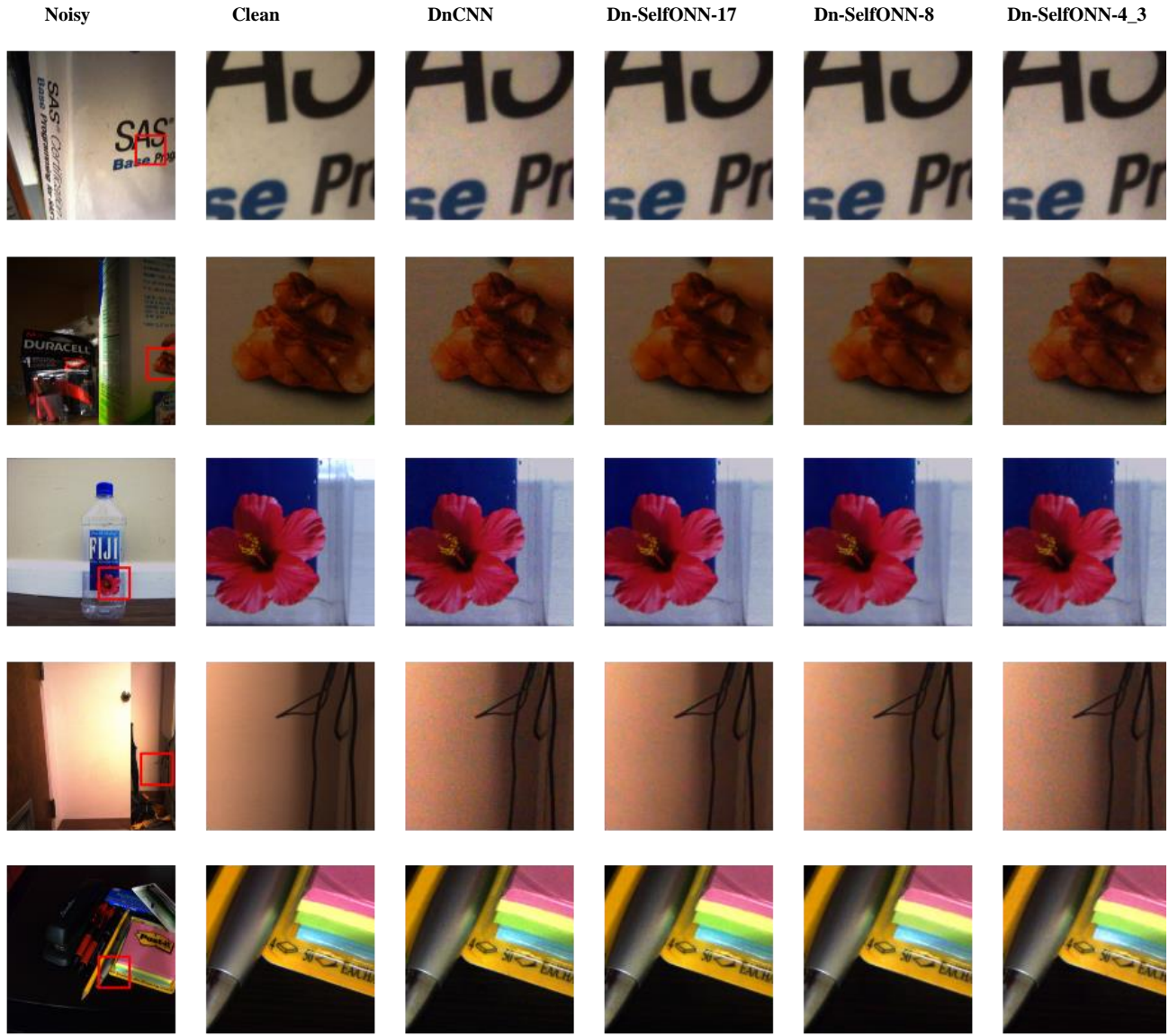

*Figure 3. Visual comparison of denoising results on high-resolution noisy images from the RENOIR [13] dataset.*

3) *Comparison on DND public benchmark*

We also evaluated the performance of DnCNN along with the Self-ONN variants on the DND benchmark dataset using their online submission system. The results of the networks used in this study and some of the well-known denoising methods are presented in Table 3. Dn-SelfONN-17 surpasses the baseline CNN by a significant margin of 0.38dB while the 8-layer Self-ONN marginally surpasses the baseline CNN. Also, the best performing 4-layer network, Dn-SelfONN-4_5 results in a decrease of only 1.2% PSNR with 4 times fewer layers and a 44.5% decrease in the number of trainable parameters.

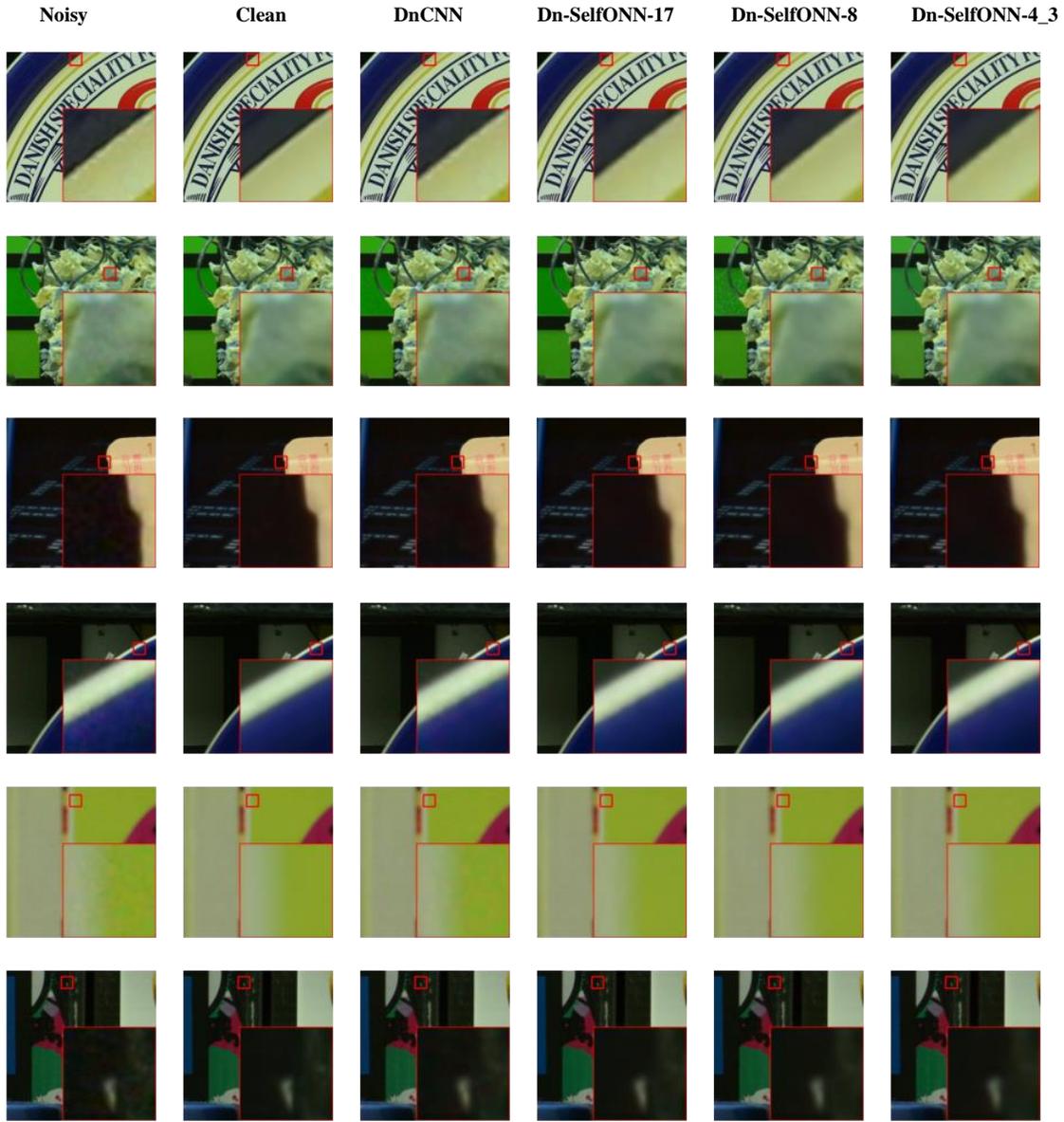

*Figure 4. Denoising results of the networks used in this study on example images from the CCNoise [22] dataset.*

Another important observation from Table 3 is that the CNN baseline used in this study, DnCNN, surpasses the method of CBDNet [26] despite the latter employing a highly enriched mixture of real and synthetic noisy images and significantly more sophisticated network architecture. Although beyond the scope of this study, this nevertheless points out a pertinent need for a fairer benchmarking of new architectures and the importance of revisiting older ones with better training data.

*Table 3. Results on DND public benchmark.*

| Network | PSNR | SSIM |
|---|---|---|
| Noisy | 29.84 | 0.7018 |
| BM3D [23] | 34.51 | 0.8507 |
| TNRD [35] | 33.65 | 0.8306 |
| KSVD [36] | 36.49 | 0.8978 |
| FFDNet+ [37] | 37.61 | 0.9415 |
| DnCNN+ | 37.90 | 0.9430 |
| TWSC [38] | 37.94 | 0.9416 |
| CBDNet [26] | 38.06 | 0.9421 |
| DnCNN (Baseline) | 38.10 | 0.9323 |
| Dn-SelfONN-17 | 38.48 | 0.9404 |
| Dn-SelfONN-8 | 38.12 | 0.9356 |
| Dn-SelfONN-4_3 | 37.56 | 0.9288 |
| Dn-SelfONN-4_5 | 37.62 | 0.9283 |

*D. Non-linearity Analysis*

*1) Inter-layer differences in non-linearity*

In Figure 5, we analyze the distribution of learned non-linearities between the layers of the trained network. For representing the power of weights, we simply use the variance as a direct measure of the relative impact or synaptic strength of each order of weights. This has been found useful in earlier studies dealing with operator optimization of ONNs [19]. Recall from Section III, that Q=1 corresponds to the weights of the convolutional neuron, whereas weights for Q>1 are responsible for encoding non-linearity into the transformation. Therefore, we plot the variance of subsets of weights corresponding to different Q values. From Figure 5, we immediately notice clear trends that are consistent across different network topologies. The input layer seems to always favor the linear convolution model as all three Self-ONN networks, Dn-SelfONN-17, Dn-SelfONN-8, and Dn-SelfONN-4 have the highest variance for Q=1 weights for the input layer. Interestingly, the opposite is observed for the output layer. Specifically, all three networks tend to exclusively favor the higher-order weights, pointing towards a greater need for non-linearity. As outlined in earlier works [16,20], this is one of the key benefits of Self-ONN. As compared to ONNs where the output layer operator is fixed, Self-ONNs can generate the required non-linear transformation for the output layer like any other layer in the network. Overall, the tendency of denoising Self-ONN networks to incline towards a linear (convolutional) model for the input layer and a highly non-linear transformation for the output layer provides evidence that the naïve approach of using the same Q value for all layers is sub-optimal and a better approach would be to use a lower value of Q for layers closer to the input and a higher value only for the last few layers. This will also have an additional benefit of decreasing the total number of trainable parameters, especially in the case of deep networks

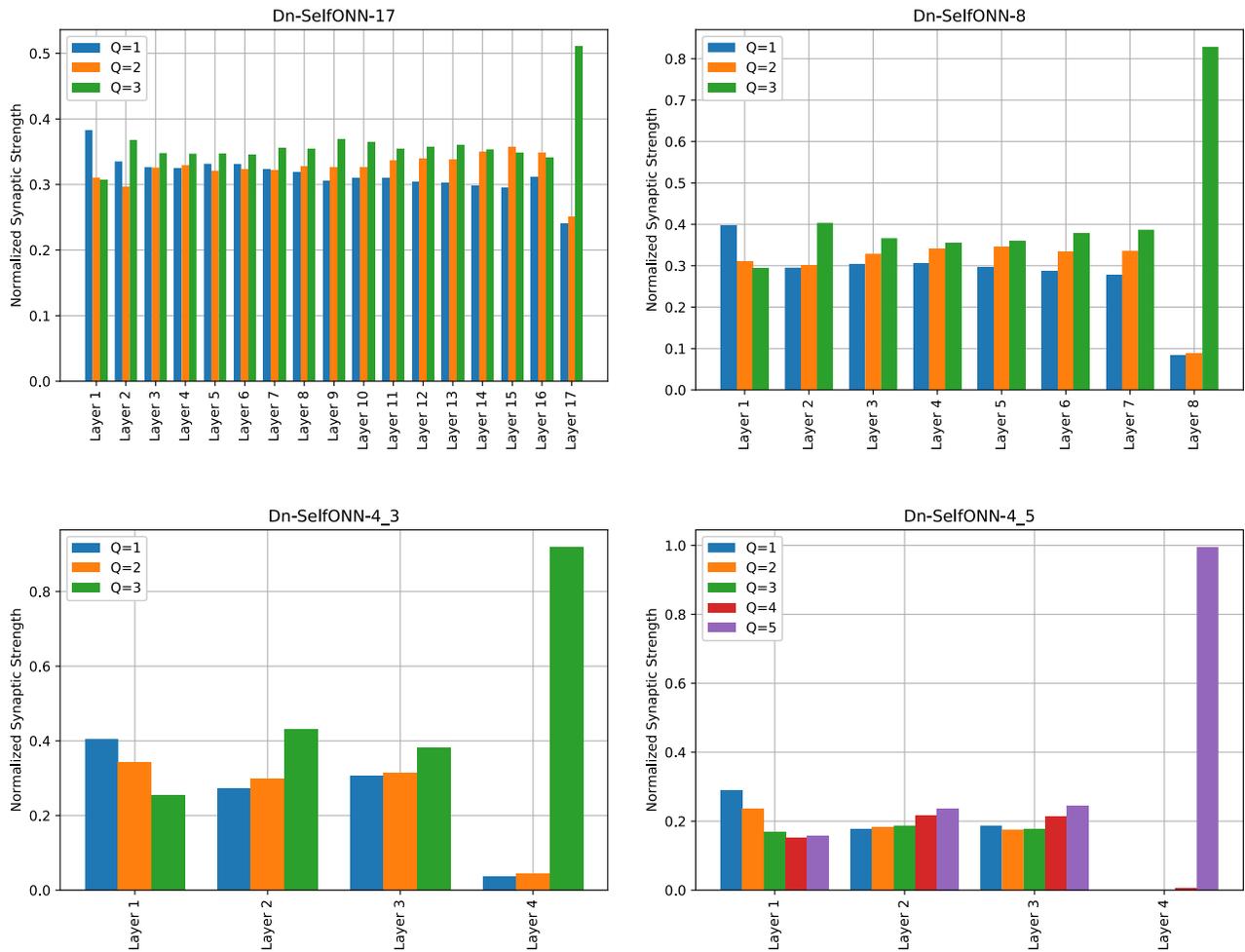

*Figure 5. Distribution of weights corresponding to different Q values in the Self-ONN networks used in this study.*

*2) Effect of depth on non-linearity*

Earlier works [16,20] have shown how ONNs and Self-ONNs, can learn sufficiently well with compact networks and surpass the linear model of CNNs. This ability is often attributed to the fact that Self-ONNs can incorporate higher non-linearity even in small network architectures. As this is the first study dealing with deep Self-ONNs, it is worth analyzing how the trained deep networks make use of the available non-linear transformations and differ from the corresponding compact networks trained for the same problem. To assess this, we measure the synaptic strength of connections corresponding to different Q values for Self-ONN networks, as shown in Figure 6. For the deepest network, Dn-SelfONN-17, the strengths corresponding to the three Q values are distributed approximately evenly with Q>1 weights slightly overpowering Q=1 weights. For Dn-SelfONN-8, we start noticing the need for higher non-linearities as Q=3 connections are stronger compared to Q=2 and the convolutional (Q=1) connections. Finally, for the compact 4-layer network, the need for non-linearity is greatly increased, reflected by more than the 2-times increase in the synaptic strength from Q=2 to Q=3, and more than the 4-times increase for the case of convolutional (Q=1) connections. This is another evidence towards the premise that the convolutional neuron model is inherently suited for deeper networks and is not expressive enough to learn meaningful representations in the compact domain.

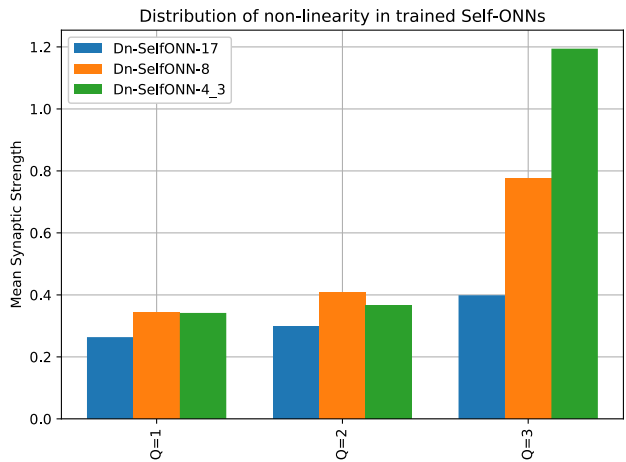

*Figure 6. The extent of learned non-linearity for the Self-ONN networks with different depths.*

E. *Computational Complexity Comparison*

Multiply and Accumulate (MAC) operations of Self-ONNs were estimated in [ref where Self-ONN MAC operations were computed]. In this section, we present a comprehensive comparison of the computational complexity of both Self-ONNs and CNNs by estimating the total number of FLOPs for both networks. Afterward, we provide a discussion regarding the parallelizability of the operations and its impact on the inference times of the networks. Without loss of generality, we assume the 'same' convolution with unit stride and zero dilation, so the layers' input and output spatial resolution are kept identical. As none of the networks used in this study use bias, we omit that in our formulation.

Given a layer $l$ in the network architecture, we denote the number of neurons in the current layer (output channels) as $N_l$ and the number of neurons in the previous layer (input channels) as $N_{l-1}$. For the first (input) layer, $N_{l-1}$ corresponds to the number of channels in the input image (3 for this study). The kernel size is assumed to be $K \times K$ and the spatial resolution of the input (and output) is $H \times W$. The number of FLOPs required for the CNN layer is now given by the following formula:

$$FLOPs_{CNN}^l = (H * W)(K^2 N_{l-1} N_l + (K^2 - 1) N_{l-1} N_l) \qquad (28)$$

The terms $K^2 N_{l-1} N_l$ and $(K^2 - 1) N_{l-1} N_l$ represent the total number of multiplication and addition operations required for the kernel operations of this layer, respectively. As each pixel in the output requires kernel operations from all the neurons, the above terms are multiplied by the number of pixels in the output. This can be visually verified from (15). Now, for an operational layer, we can use the same notation. Note from (27) that the Self-ONN formulation can be represented as a convolution operation with Q-times more input channels than the CNN layer, albeit with an additional cost of achieving the Hadamard exponentiation operation of (20). Taking this into account, we can write the total cost of a Self-ONN layer in terms of the number of FLOPs as follows:

$$FLOPs_{SelfONN}^l = \left((H * W)(K^2 N_{l-1} Q N_l + (K^2 - 1) N_{l-1} Q N_l)\right) + FLOPs_{exp}^Q \qquad (29)$$

where $FLOPS_{exp}^Q$ are the multiplication operations required to achieve the Hadamard exponentiation and are calculated as follows:

$$FLOPs_{exp}^Q = H * W * N_{l-1} * T(Q-1) \qquad (30)$$

In (30), $T(Q-1)$ denotes the $(Q-1)^{th}$ triangular number and is calculated as $\sum_{k=1}^{Q-1} k$. Note that this assumes that all higher powers of the input are calculated independently from each other and do not reuse the previous power i.e. $x^q$ is calculated as $\prod_1^q x$ instead of $x^{q-1} x$. Despite being apparently redundant, this is important to preserve the independence of the operations because reusing previous powers entails waiting for all $q-1$ powers before finally calculating $x^q$. In practice, this will enable efficient exploitation of parallel computing resources such as CUDA cores.

Finally, using (30), the number of FLOPS required for a Self-ONN layer can be calculated. Table 4 provides the number of FLOPS for all networks used in this study. Note that the computational cost of activation functions and BatchNorm layers are omitted as they have the same parameters for both CNN and Self-ONN layers and roughly add a constant number of FLOPS. It is key to note here that the total number of FLOPS, despite wide-scale usage, lacks a definitive factor that ultimately determines the efficiency of the network - the parallelizability factor. Specifically, the ratio of operations in the network which are independent of each other, and hence can be parallelized in an ideal setting, is not accounted for. Therefore, in practice, FLOPS alone cannot be used as a sole measure of network efficiency. For example, given the availability of enough parallel computing resources, a model requiring $F$ number of FLOPs (distributed evenly among the layers), composed of 10 layers can only achieve $F/10$ FLOPS in a unit time because each intermediate layer is dependent on the output of the previous layer. However, for a 2-layer network with the same number of FLOPs, $F/2$ FLOPS can be calculated in a unit time, resulting in a 5-fold theoretical speed-up as compared to the 10-layer network. Therefore, the shallower denoising networks used in this study, Dn-SelfONN-8 and Dn-SelfONN-4 lend themselves to more efficient parallelization. Even though the current Self-ONN implementation is not optimized to fully exploit the independence of operations, we can see the above effect reflected quite clearly in the empirical observations of the running times of the different networks in Table 4. For example, we see that Dn-SelfONN-4_5, despite having 67% more FLOPs compared to Dn-SelfONN-4_3, results in an increase of only 43% in terms of running time. More importantly, Dn-SelfONN-8 with 8-layers, despite requiring 21% more FLOPs compared to the 17-layer CNN, achieves more than 8% decrease in running time. In general, the running times of the GPU-powered Self-ONNs can be further reduced if a fused kernel approach is adopted, such that the Hadamard exponentiation operation and the convolution are carried out by the same CUDA kernel.

*Table 4. Number of FLOPs and the running times for the denoising networks used in this study.*

|  | FLOPs (G) | Running Time (ms) | |
| --- | --- | --- | --- |
|  |  | 256x256 | 512x512 |
| **DnCNN** | 68.88 | 11.720 | 41.888 |
| **Dn-SelfONN-17** | 207.04 | 24.757 | 92.925 |
| **Dn-SelfONN-8** | 83.60 | 10.691 | 39.539 |
| **Dn-SelfONN-4_3** | **28.74** | **4.433** | **15.789** |
| **Dn-SelfONN-4_5** | 47.96 | 6.366 | 23.389 |

## V. CONCLUSION

We propose a novel deep Self-ONN for real-world image denoising problems. Over the four benchmark datasets, this study has demonstrated for the first time that the resulting Self-ONNs achieve a superior denoising performance than the competing deep CNN architecture, DnCNN. Moreover, a similar or even better performance level can still be achieved by Self-ONNs using a fraction of the layers and neurons of the DnCNN. This in fact questions the need for creating very deep and complex networks since this study has clearly shown that the required diversity and learning/generalization capability can also be accomplished by the proposed self-organized "heterogeneous" networks with compact configurations both in terms of network depth and the total number of learning units (neurons). Furthermore, this study has revealed that both in deep and compact networks, the need for non-linearity is higher for layers closer to the output layer compared to layers closer to the input layer. The study further revealed that for the denoising problem, compact Self-ONNs require more non-linearity (higher Q) to stay competitive with deeper networks. Optimizing the level of non-linearity per layer remains a key challenge towards further improving the performance of Self-ONNs and reducing their network complexity. Moreover, the proposed formulation of Self-ONN enables more efficient exploitation of the independent operations, resulting in a better running time as compared to the CNN, despite requiring more FLOPs. Unsupervised adaptation of layer-wise non-linearity and a more efficient implementation remain open topics to be tackled in future extensions of this work.

## VI. REFERENCES

[1] J. Portilla, V. Strela, M.J. Wainwright, E.P. Simoncelli, Image denoising using scale mixtures of Gaussians in the wavelet domain, IEEE Transactions on Image Processing. (2003). doi:10.1109/TIP.2003.818640.

[2] I. Tăbuş, D. Petrescu, M. Gabbouj, A training framework for stack and Boolean filtering - Fast optimal design procedures and robustness case study, IEEE Transactions on Image Processing. (1996). doi:10.1109/83.503901.

[3] H.C. Burger, C.J. Schuler, S. Harmeling, Image denoising: Can plain neural networks compete with BM3D?, in: Proceedings of the IEEE Computer Society Conference on Computer Vision and Pattern Recognition, 2012. doi:10.1109/CVPR.2012.6247952.

[4] K. Zhang, W. Zuo, Y. Chen, D. Meng, L. Zhang, Beyond a Gaussian denoiser: Residual learning of deep CNN for image denoising, IEEE Transactions on Image Processing. (2017). doi:10.1109/TIP.2017.2662206.

[5] V. Singhal, A. Majumdar, A domain adaptation approach to solve inverse problems in imaging via coupled deep dictionary learning, Pattern Recognition. (2020). doi:10.1016/j.patcog.2019.107163.

[6] I. Hong, Y. Hwang, D. Kim, Efficient deep learning of image denoising using patch complexity local divide and deep conquer, Pattern Recognition. (2019). doi:10.1016/j.patcog.2019.06.011.

[7] Y. Quan, Y. Chen, Y. Shao, H. Teng, Y. Xu, H. Ji, Image denoising using complex-valued deep CNN, Pattern Recognition. (2021). doi:10.1016/j.patcog.2020.107639.

[8] T. Plötz, S. Roth, Benchmarking denoising algorithms with real photographs, in: Proceedings - 30th IEEE Conference on Computer Vision and Pattern Recognition, CVPR 2017, 2017. doi:10.1109/CVPR.2017.294.

[9]  A. Foi, M. Trimeche, V. Katkovnik, K. Egiazarian, Practical Poissonian-Gaussian noise modeling and fitting for single-image raw-data, IEEE Transactions on Image Processing. (2008). doi:10.1109/TIP.2008.2001399.

[10] F. Luisier, T. Blu, M. Unser, Image denoising in mixed poissongaussian noise, IEEE Transactions on Image Processing. (2011). doi:10.1109/TIP.2010.2073477.

[11] Y. Zhou, J. Jiao, H. Huang, Y. Wang, J. Wang, H. Shi, T. Huang, When AWGN-based Denoiser Meets Real Noises, (2019). http://arxiv.org/abs/1904.03485 (accessed April 1, 2020).

[12] A. Abdelhamed, S. Lin, M.S. Brown, A High-Quality Denoising Dataset for Smartphone Cameras, in: Proceedings of the IEEE Computer Society Conference on Computer Vision and Pattern Recognition, 2018. doi:10.1109/CVPR.2018.00182.

[13] J. Anaya, A. Barbu, RENOIR – A dataset for real low-light image noise reduction, Journal of Visual Communication and Image Representation. (2018). doi:10.1016/j.jvcir.2018.01.012.

[14] C. Tian, L. Fei, W. Zheng, Y. Xu, W. Zuo, C.W. Lin, Deep learning on image denoising: An overview, Neural Networks. (2020). doi:10.1016/j.neunet.2020.07.025.

[15] S. Kiranyaz, T. Ince, A. Iosifidis, M. Gabbouj, Operational neural networks, Neural Computing and Applications. (2020). doi:10.1007/s00521-020-04780-3.

[16] J. Malik, S. Kiranyaz, M. Gabbouj, Self-organized operational neural networks for severe image restoration problems, Neural Networks. (2020). doi:https://doi.org/10.1016/j.neunet.2020.12.014.

[17] C. Wang, J. Yang, L. Xie, J. Yuan, Kervolutional neural networks, in: Proceedings of the IEEE Computer Society Conference on Computer Vision and Pattern Recognition, 2019. doi:10.1109/CVPR.2019.00012.

[18] G. Zoumpourlis, A. Doumanoglou, N. Vretos, P. Daras, Non-linear Convolution Filters for CNN-Based Learning, in: Proceedings of the IEEE International Conference on Computer Vision, 2017. doi:10.1109/ICCV.2017.510.

[19] S. Kiranyaz, J. Malik, H.B. Abdallah, T. Ince, A. Iosifidis, M. Gabbouj, Exploiting Heterogeneity in Operational Neural Networks by Synaptic Plasticity, Neural Computing and Applications. November (2020).

[20] S. Kiranyaz, J. Malik, H. Ben Abdallah, T. Ince, A. Iosifidis, M. Gabbouj, Self-Organized Operational Neural Networks with Generative Neurons, Neural Networks. (2020). doi:https://doi.org/10.1016/j.neunet.2021.02.028.

[21] J. Malik, S. Kiranyaz, M. Yamac, M. Gabbouj, BM3D vs 2-Layer ONN, (2021). http://arxiv.org/abs/2103.03060 (accessed April 8, 2021).

[22] S. Nam, Y. Hwang, Y. Matsushita, S.J. Kim, A Holistic Approach to Cross-Channel Image Noise Modeling and Its

[22] Application to Image Denoising, in: Proceedings of the IEEE Computer Society Conference on Computer Vision and Pattern Recognition, 2016. doi:10.1109/CVPR.2016.186.

[23] K. Dabov, A. Foi, V. Katkovnik, K. Egiazarian, Image restoration by sparse 3D transform-domain collaborative filtering, in: Image Processing: Algorithms and Systems VI, 2008. doi:10.1117/12.766355.

[24] S.W. Zamir, A. Arora, S. Khan, M. Hayat, F.S. Khan, M.H. Yang, L. Shao, Learning Enriched Features for Real Image Restoration and Enhancement, in: Lecture Notes in Computer Science (Including Subseries Lecture Notes in Artificial Intelligence and Lecture Notes in Bioinformatics), 2020. doi:10.1007/978-3-030-58595-2_30.

[25] S. Anwar, N. Barnes, Real image denoising with feature attention, in: Proceedings of the IEEE International Conference on Computer Vision, 2019. doi:10.1109/ICCV.2019.00325.

[26] S. Guo, Z. Yan, K. Zhang, W. Zuo, L. Zhang, Toward convolutional blind denoising of real photographs, in: Proceedings of the IEEE Computer Society Conference on Computer Vision and Pattern Recognition, 2019. doi:10.1109/CVPR.2019.00181.

[27] C. Chen, Z. Xiong, X. Tian, Z.J. Zha, F. Wu, Real-World Image Denoising with Deep Boosting, IEEE Transactions on Pattern Analysis and Machine Intelligence. (2020). doi:10.1109/TPAMI.2019.2921548.

[28] D.W. Kim, J.R. Chung, S.W. Jung, GRDN:Grouped residual dense network for real image denoising and GAN-based real-world noise modeling, in: IEEE Computer Society Conference on Computer Vision and Pattern Recognition Workshops, 2019. doi:10.1109/CVPRW.2019.00261.

[29] S.W. Zamir, A. Arora, S. Khan, M. Hayat, F.S. Khan, M.H. Yang, L. Shao, CycleISP: Real image restoration via improved data synthesis, in: Proceedings of the IEEE Computer Society Conference on Computer Vision and Pattern Recognition, 2020. doi:10.1109/CVPR42600.2020.00277.

[30] Y. Song, Y. Zhu, X. Du, Grouped Multi-Scale Network for Real-World Image Denoising, IEEE Signal Processing Letters. (2020). doi:10.1109/LSP.2020.3039726.

[31] Y. Quan, M. Chen, T. Pang, H. Ji, Self2self with dropout: Learning self-supervised denoising from single image, in: Proceedings of the IEEE Computer Society Conference on Computer Vision and Pattern Recognition, 2020. doi:10.1109/CVPR42600.2020.00196.

[32] Y. Zhao, Z. Jiang, A. Men, G. Ju, Pyramid Real Image Denoising Network, in: 2019 IEEE International Conference on Visual Communications and Image Processing, VCIP 2019, 2019. doi:10.1109/VCIP47243.2019.8965754.

[33] L. Bao, Z. Yang, S. Wang, D. Bai, J. Lee, Real image denoising based on multi-scale residual dense block and cascaded

U-net with block-connection, in: IEEE Computer Society Conference on Computer Vision and Pattern Recognition Workshops, 2020. doi:10.1109/CVPRW50498.2020.00232.

[34]   J. Malik, S. Kiranyaz, M. Gabbouj, FastONN – Python based open-source GPU implementation for Operational Neural Networks, ArXiv. (2020).

[35]   Y. Chen, T. Pock, Trainable Nonlinear Reaction Diffusion: A Flexible Framework for Fast and Effective Image Restoration, IEEE Transactions on Pattern Analysis and Machine Intelligence. (2017). doi:10.1109/TPAMI.2016.2596743.

[36]   M. Aharon, M. Elad, A. Bruckstein, K-SVD: An algorithm for designing overcomplete dictionaries for sparse representation, IEEE Transactions on Signal Processing. (2006). doi:10.1109/TSP.2006.881199.

[37]   K. Zhang, W. Zuo, L. Zhang, FFDNet: Toward a fast and flexible solution for CNN-Based image denoising, IEEE Transactions on Image Processing. (2018). doi:10.1109/TIP.2018.2839891.

[38]   J. Xu, L. Zhang, D. Zhang, A trilateral weighted sparse coding scheme for real-world image denoising, in: Lecture Notes in Computer Science (Including Subseries Lecture Notes in Artificial Intelligence and Lecture Notes in Bioinformatics), 2018. doi:10.1007/978-3-030-01237-3_2.